**Evolving choice hysteresis in reinforcement learning: comparing the adaptive value of positivity bias and gradual perseveration**

Isabelle Hoxha[1,2], Léo Sperber[1,2], Stefano Palminteri[1,2]

1) Département d'études cognitives, École normale supérieure, Université de Recherche Paris Science et Lettres, Paris, France
2) Laboratoire de Neurosciences Cognitives et computationnelles, Institut National de la Santé et de la Recherche Médicale, Paris, France

**Abstract**
The tendency of repeating past choices more often than expected from the history of outcomes has been repeatedly empirically observed in reinforcement learning experiments. It can be explained by at least two computational processes: asymmetric update and (gradual) choice perseveration. A recent meta-analysis showed that both mechanisms are detectable in human reinforcement learning. However, while their descriptive value seems to be well established, they have not been compared regarding their possible adaptive value. In this study, we address this gap by simulating reinforcement learning agents in a variety of environments with a new variant of an evolutionary algorithm. Our results show that positivity bias (in the form of asymmetric update) is evolutionary stable in many situations, while the emergence of gradual perseveration is less systematic and robust. Overall, our results illustrate that biases can be adaptive and selected by evolution, in an environment-specific manner.

## Introduction

In recent years, empirical studies of reinforcement learning in humans and other species have consistently shown that agents tend to repeat their choices more than what is expected based on standard computational cognitive models[1–5]. In other terms, experimental subjects then repeat their choice more than what is accounted by an objective an unbiased representation of the history of past outcomes. This phenomenon, known as choice repetition (positive autocorrelation or hysteresis), has been found to be present in a wide range of tasks, including two-armed bandit tasks, and has important implications for our understanding of how agents make decisions from experience.

There are two main computational processes that have been proposed to account for choice repetition in the context of reinforcement learning. The first is an outcome-based process, which consists in an asymmetric update of estimated values, whereby the learning rate is higher for positive outcomes compared to negative outcomes[6], a processes often referred to as positivity bias or optimistic update. The second is a choice-based process, which involves repeating previous choices regardless of the previous outcome, known as gradual perseveration[7]. The former (asymmetric update) is thought to represent the reinforcement learning counterpart of the otherwise pervasive confirmation and self-serving biases[8,9]. The latter (gradual perseveration) is often explained as computational instantiation of choice-based preference change[10–12] or habit formation[13].

The debate about which of these processes provides a better account of empirical data in human experiments has been recently ecumenically settled by a meta-analysis finding evidence for both processes in 9 different datasets[2,3]. Thus, both asymmetric update and gradual perseveration seem to be both important as *descriptive* features of human reinforcement learning. However, their *normative* status has not been systematically assessed and compared yet. In fact, while some studies have investigated the optimality of asymmetric update, they have focused either on quite specific environments[14], or feedback information regimes[15] and they have not compared it with gradual perseveration. The optimality of gradual perseveration has seldom been investigated, and in any case not in bandit tasks[16,17].

In this paper, we address this question by deriving a new variant of an evolutionary algorithm and simulating agents in several variants of two-armed bandit tasks, similar to those used in the lab to empirically study reinforcement learning in humans and animals. In addition to the crucial inclusion of gradual perseveration and the utilization of evolutionary simulations, our analysis differs from previous ones by our focus on partial feedback regimen and considering a broad range of task's outcome contingencies[14]. More specifically, across simulations we manipulated the task difficulty (difference in value between options), richness (the average value of the two options), how many times a given pair of options was presented and, finally, task volatility, in terms of reversal frequency and probability distributions.

Evaluating and understanding the adaptive value of positivity bias and gradual perseveration present several key advantages. First, it can shed light on an apparent puzzle by helping us understand why such biases have not been eliminated by natural selection. In doing so, our study contributes to the greater debate concerning the discrepancy between statistical optimality and human bounded rationality[18–22]. Second, our results can also inform artificial intelligence by identifying situations where, somehow counterintuitively, biased agents outperform unbiased ones[23–25].

# Results

## General structure of the task and computational model

Our reinforcement learning agents performed two-armed bandit tasks, whereby they had to learn which of two alternatives (referred to as $A$ and $B$) yield the highest average reward while receiving stochastic feedback only for the chosen option (**Figure 1A**). We devised different variants of this task (referred to as 'environments'), where we manipulated the process generating the rewards (**Figure 1B**; the details concerning the variants will be provided later when introducing the relevant results). Our agents were represented by a simple computational model illustrated in **Figure 1C**. The agents decided which action to take from a probability defined (for option $A$) as

$$P_A = \frac{1}{1 + exp(-\beta(Q_A - Q_B) - \phi(C_A - C_B))}$$

where $\beta$ (choice inverse temperature) and $\phi$ (choice trace bias) are the parameters that weight the relative contributions of the Q-values and choice traces (C-values). Q-values are option-specific latent variables that store information about past outcomes. The Q-value of the chosen option ($a$) is updated at each outcome step following a delta (or Rescorla-Wagner) rule:

$$Q_c \leftarrow Q_c + \begin{cases} \alpha_+ \times PE_c & (if\ PE_c > 0) \\ \alpha_- \times PE_c & (if\ PE_c < 0) \end{cases},$$

where $PE_c$ is the reward prediction error is standardly computed as

$$PE_c = r_c - Q_{s,c},$$

The learning rates $\alpha_+$ and $\alpha_-$ apply upon positive and negative feedback, respectively. Our agents therefore allow for asymmetric update, whereby the impact of positive and negative prediction errors is different ($\alpha_+ \neq \alpha_-$). C-values are also option-specific latent variables, but they store information concerning the past frequency of choice. After a decision is made, the C-value of the chosen options is updated as follows:

$$C_c \leftarrow C_c + \tau \times (1 - C_c)$$

While that of the unchosen option ($u$) decays:

$$C_u = (1 - \tau) \times C_u$$

The parameter $\tau$ is called the choice trace learning rate and determine the speed of the update of the C-values.

An agent $i$ is therefore characterized by a vector of five parameters:

$$\vec{a_i} = [a_1, ... a_n] = [\alpha_+, \alpha_-, \beta, \tau, \phi].$$

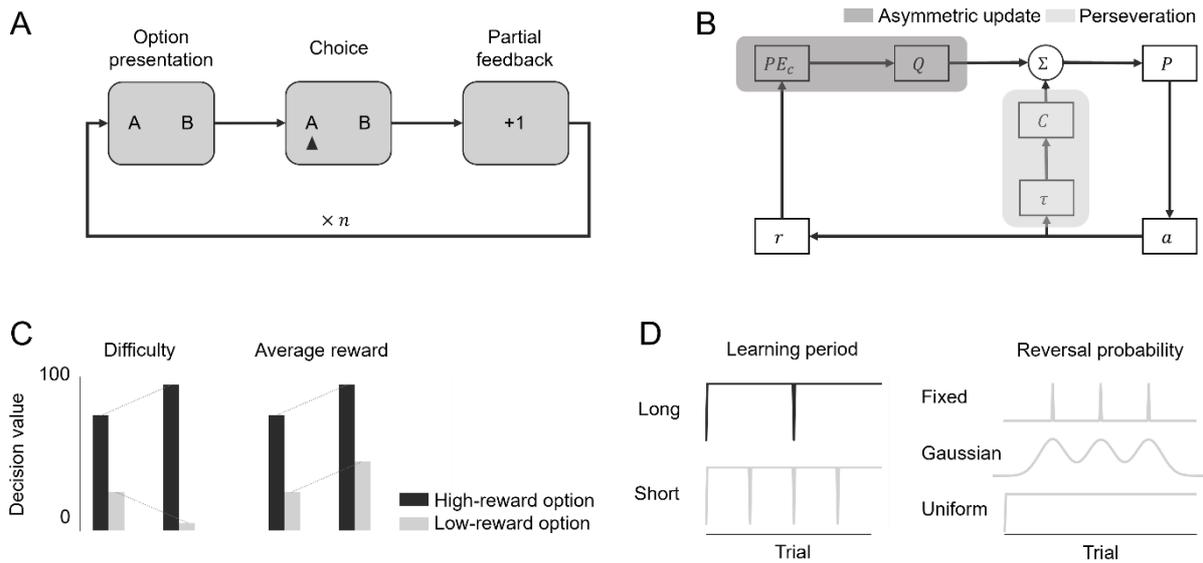

**Figure 1**: the task and the manipulated factors. (**A**) the bandit task: at each trial, a choice is given to the virtual agents, with different reward probabilities. Once the choice is made, the agents are given feedback on the chosen option. (**B**) the learning process: agents learn from previous rewards and previous choices. We therefore introduce two biases: a value bias (positive and negative feedback are allowed to update the choice rule differently), and a perseveration bias (agents are likely to repeat or alternate choices). (**C**) four factors are manipulated. The difficulty, whereby greater differences between reward probability correspond to lower difficulty, the average reward, whereby the difference between reward probabilities remain constant but the expected average value increases, the learning period, in which the state is changed more or less frequently, and the volatility, in which reversal of values is introduced at each change of state with different frequencies and probability functions. In the fixed condition, the state changes periodically after a fixed number of trials. In the Gaussian condition, the state changes at a moment around the fixed reversal state changes with a Gaussian distribution. In the Uniform condition, the state can change at any time with a fixed probability $p = \frac{n_{reversals}}{n_{trials}}$, with $n_{trials}$ the number of trials over the simulation and $n_{reversals}$ the number of reversals occurring in the corresponding fixed reversal simulation.

### The evolutionary algorithm

We searched for optimal sets of parameters by submitting our artificial agents to an evolutionary selection algorithm, where their "genotype" is represented by the values in the parameters' vector (in other terms the genotype has five *loci*). In our algorithm, the relevant "phenotype" was represented by the behavior of our agents in the bandit task and the fitness was proportional to the average rate of reward-maximizing after 160 trials (the number of trials has been chosen to be similar to those used in human experimental studies[3]). More precisely, what determined reproductive success (and therefore the transmission of a genome to the next generation) was the relative rank in terms of fitness (absolute measures of performance could not be used, since the environments we explored in our simulations radically differed in difficulty levels and their average final accuracy).

In each evolutionary step the population number was $n_{agents} = 1000$. After each generation, following the principle of the 'survival of the fittest', the agents ranked at the bottom 5% ($n_{agents} = 50$) of accuracy did not transmit their genome to the next generation, while the top 95% agent did not ($n_{agents} = 950$). A further bonus is then given to the top 5% ($n_{agents} = 50$) agents which give two (instead of one) "descendants" to the next generation (i.e. their genome is transmitted to two new agents). The bonus given to the individuals with highest fitness (in addition of been biologically plausible) has the advantage to ensure that the number of agents remain constant ($n_{agents} = 1000$) and to accelerate the discovery of good genotypes (as shown in preliminary tests of the evolutionary algorithm) (**Figure 2A**).

In our study we then considered two possible ways to introduce genotypic variability (**Figure 2B**). A first (biologically plausible) possibility is to introduce random mutations as small numerical changes (5% of their possible ranges) in the parameters value of some agents. To speed up the evolution process, we implemented random mutations in the descendance of the 95% remaining agents. To avoid issues such as an incomplete exploration of the parameter space, falling to local minima of the fitness function and arbitrary choice of initial parameters, we considered another (admittedly less biological plausible) way to introduce genetic variability, by starting with higher variable individual whose parameter values were randomly drawn from their possible ranges.

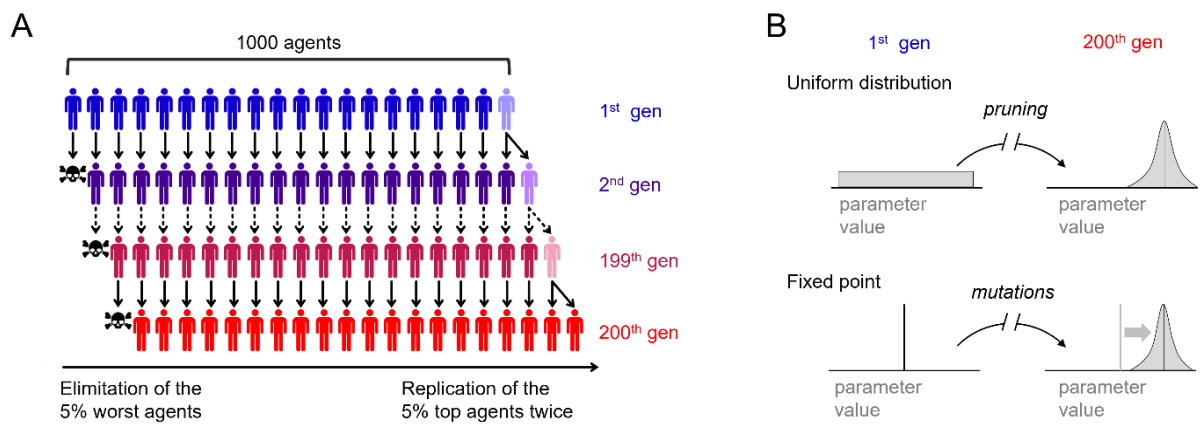

**Figure 2**: The evolutionary algo (**A**) Schematic illustrating the evolutionary process (1000 agent were simulated for 200 generations). Each agent performs all trials, and its average accuracy is computed. The 5% of agents with the lower accuracy are sacrificed, while the top 5% are duplicated. (**B**) Schematic illustrating the consequence of the selection process on the parameters' distributions, given two scenarios of initialization. In the first case (top row), all agents are initialized with the same parameters, and random mutations across generations lead to a spreading of the parameter distribution. In the second case (bottom row), agents' parameters are sampled from a uniform distribution, and the selection process narrows down the parameter distributions.

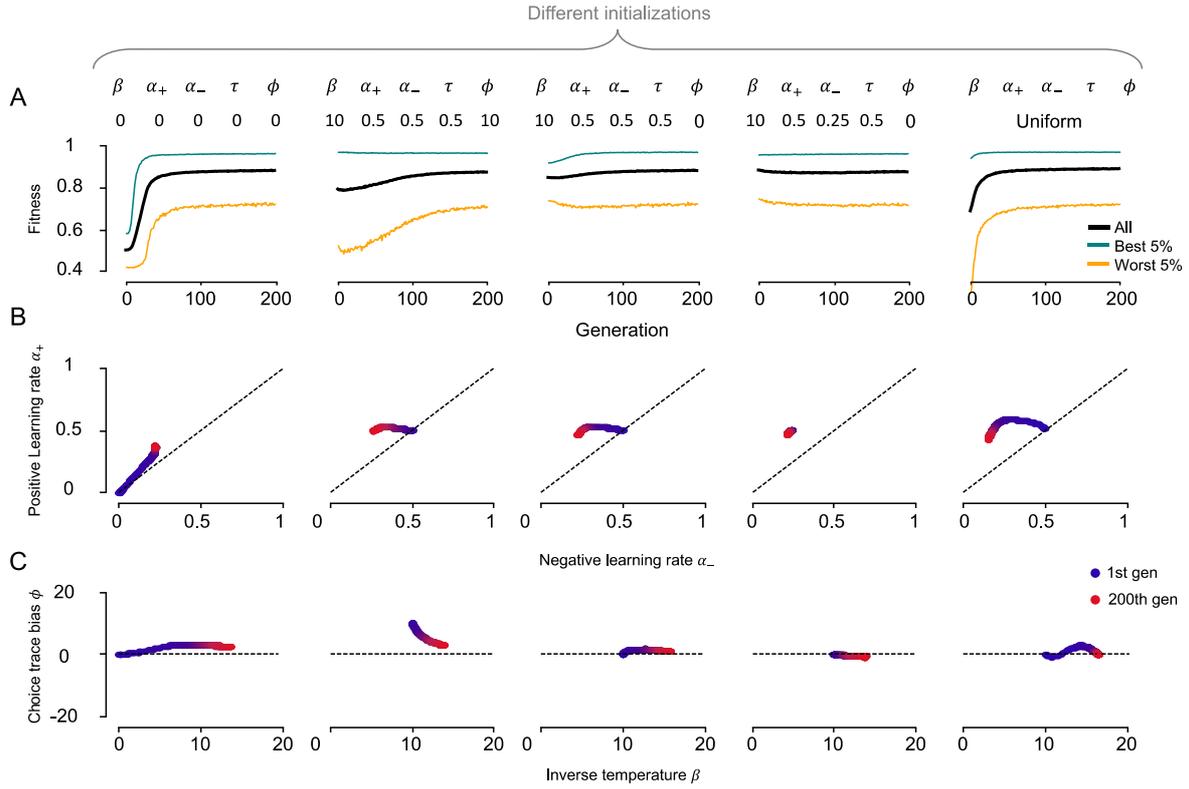

**Figure 3 stability of the results across different initializations** The results correspond to simulations with difficulty = 75/25 and learning period = 20(8)[1,26]. (**A**) The top row represents the average fitness of agents across generations, for the top 5% agents (blue), the worst 5% (yellow), and the average over all agents (black). (**B**) The middle row represents the population average learning rate for positive outcomes against the learning rate for negative outcomes across generations. Point above/below the diagonal corresponds to a positivity/negativity bias. (**C**) The bottom row represents the population average choice trace weight against the temperature across generations. Points above/below zero correspond to a tendency to repeat/change from previous choices. The blue-to-red color gradient of the dots is proportional to the generation (first: blue, last: red).

**Table 1:** predominance of biases across different initializations. Each initialization was simulated 100 times (each simulation involving 1000 agents and 200 generations). Each entry denotes the percentage of reboots presenting with type of bias and its directions ($\phi > 0$: perseveration, $\alpha_+ > \alpha_-$: positivity bias).

| Initial parameters' values | $\phi > 0$ | $\alpha_+ > \alpha_-$ |
|---|---|---|
| $\beta = \alpha_+ = \alpha_- = \tau = \phi = 0$ | 100% | 100% |
| $\beta = 10, \alpha_+ = \alpha_- = \tau = 0.5, \phi = 10$ | 100% | 100% |
| $\beta = 10, \alpha_+ = \alpha_- = \tau = 0.5, \phi = 0$ | 64% | 100% |
| $\beta = 10, \alpha_+ = \tau = 0.5, \alpha_- = 0.25, \phi = 0$ | 0% | 100% |
| uniform | 36% | 100% |

## Evolution of parameters across different initializations

We first studied the evolution of the parameters in an environment with a task difficulty set to 25%/75% (which designate the probability of reward for the worst/best option) and 8 learning periods lasting 20 trials each. These parameters have been chosen in accordance with previous empirical investigations[3] and the results of this simulation are presented on **Figure 3**.

The top row represents the fitness across generations. We observed that the function is "well-behaved", meaning that it is a monotonic function of generation number and negatively accelerated, stabilizing at a plateau before the 200th generation to ~80% accuracy for all initializations, which ensure the correct implementation and parametrization of the evolutionary algorithm. The middle row represents the evolution of $\alpha_+$ as a function of $\alpha_-$ across generations, and the bottom row $\phi$ as a function of $\beta$ across generations.

We first focus on the case where the initial parameters are $\beta = \alpha_+ = \alpha_- = \tau = \phi = 0$. In this case, both the positivity bias ($\alpha_+ > \alpha_-$) and the gradual preservation ($\phi > 0$) are selected in 100% of the reboots of the evolution process. It is now interesting to compare to other initializations. The second column of represent a case in which the population start with gradual perseveration and no positivity bias ($\beta = 10, \alpha_+ = \alpha_- = \tau = 0.5, \phi = 10$). We observe that the positivity bias emerges, while the $\phi$ parameter is significantly reduced across generations. In the simulations in the third column, ($\beta = 10, \alpha_+ = \alpha_- = \tau = 0.5, \phi = 0$), the positivity bias is also selected in 100% of the reboots, while gradual preservation much more sporadically (64%). Finally, when the population is initialized with a positivity bias, but not gradual perseveration ($\beta = 10, \alpha_+ = 0.5, \alpha_- = 0.25, \tau = 0.5, \phi = 0$), the difference between learning rates remains constant, while the , $\phi$ never becomes positive (in fact, it becomes slightly negative in the 100% of the reboots). We also note that the case in which the population starts the positivity bias is also the case in which the initial fitness is the higher. The results presented so far indicate that, in the considered environment (relatively easy task, with medium learning periods and stable contingencies), a positivity bias is systematically selected, while gradual perseveration is less evolutionary stable and depends on the initialization of parameters (**Table 1**). In the subsequent simulations, to circumvent the initialization problem, we slightly modified our evolutionary algorithm concerning the way in which genotypical variability is introduced. More specifically, we initialized each parameter of each agent with a random value (uniformly drawn from its range) so that evolutionary process does not require any random mutations to introduce variability, but rather consists in pruning down bad phenotypes. These results are presented in the last column of **Figure 3**, and overall confirms our observations concerning the case of fixed initialization and random mutations (positivity bias is systematically selected, the gradual perseveration less so). Thus, to obviate falling in local maxima of fitness due to insufficient exploration of the parameter space, in the subsequent analyses we opted for this version of the algorithm.

**Evolution of parameters as a function of different reward environments**

We then moved to analyze the dependency of these results on the environment, we simulated the model while manipulating the values of reward parameters. More specifically we manipulated the task difficulty (i.e., the difference in reward probability between the two options), the average reward (i.e., the average level of probability of the two options), the length of the learning period and, finally, the level of volatility (see **Figures 1C and 1D**).

We first focus on the results concerning stable environments (i.e., without volatility). The difficulty was manipulated by modifying the difference between the reward probability of the two arms, which resulted in an easy (5%/95%), a medium (25%/75%) and a difficult (45%/55%) learning environment. The average reward is defined by the average of the two reward probabilities, while their difference is kept constant, i.e., at the same difficulty. This manipulation resulted in poor (5%/55%), medium (25%/75%) and rich (45%/95%) environments. By manipulating the block duration, we obtained short (32 repetitions of 5 trial bandits: 32:5), medium (8:20) and long (2:80) learning periods. Note that within these three factors, the "medium level" environment always corresponds to the one implemented previously in the previous analyses, meaning that these manipulations resulted in 7 unique environments.

Across the stable environments (**Figure 4**), we observed that the positivity bias was systematical selected by our evolutionary algorithm in all the environments and all the reboots, with only one exception: we noted that $\alpha_+$ decreases with increasing richness of the environment, while $\alpha_-$ increases, suggesting a decreasing optimism as the average expected value increases to the point that in the richest environment (45%/95%), the update asymmetry favors a pessimism bias in 64% of the reboots (**Table 2**). This results reflects the fact that is generally better to learn more from rare outcomes (i.e., in an environment in which positive outcomes are common – rich environment – the agents should pay more attention to the negative ones[14]). We also observed that reducing the learning period was associated with an increase of both the average learning rate and the positivity bias, which reflect the fact that in shorter task the agents must learn quickly and cannot wait to accumulate a lot of experience.

We also observe that, in contrast, the choice trace bias $\phi$ was generally not different from zero across reboots (see **Table 3**), except for very easy tasks (5%/95%), rich environments (45%/95%) and long learning periods (80 trials), in which case the choice trace is positive (i.e., a tendency for choice repetition evolves). Conversely, the choice trace is significantly negative for difficult environments (45%/55%) and for short learning periods (5 trials), marking a tendency to alternate in these environments. We argue that the propensity to develop perseveration under long learning periods makes the agents more robust to the stochasticity of the reward in the long run, i.e., when the agent is very likely to have identified the correct response. Conversely, in very short learning periods, correct exploration of the options is crucial and promote by negative perseveration (alternation). However, note that, even in environments where gradual perseveration was selected, the results were much less systematic (as documented on **Figure 4** and **Table 2**) underlying the less robust nature of gradual perseveration or alternation in evolution.

We now focus on the results concerning the environment with volatility. The volatility of the environment was modulated along two factors by introducing contingency reversals, following which the values of the two options were switched, so that the previously optimal response becomes the sub-optimal one and vice-versa. The distribution of the reversals (i.e., when they appear within the 160 trials), and their frequency (i.e., how many of them). More specifically, in a first set of environments, the reversals would occur at always after a fixed number of trials. In another set, we added gaussian noise around the fixed number of trials; finally, in a last set of simulations, the reversal could occur at any trial with a fixed probability. Manipulating the reversal frequency resulted in low (one reversal), medium (7 reversals), and high (31 reversals) volatility levels. These manipulations resulted in a total 9 unique volatile environments (see **Table 2**).

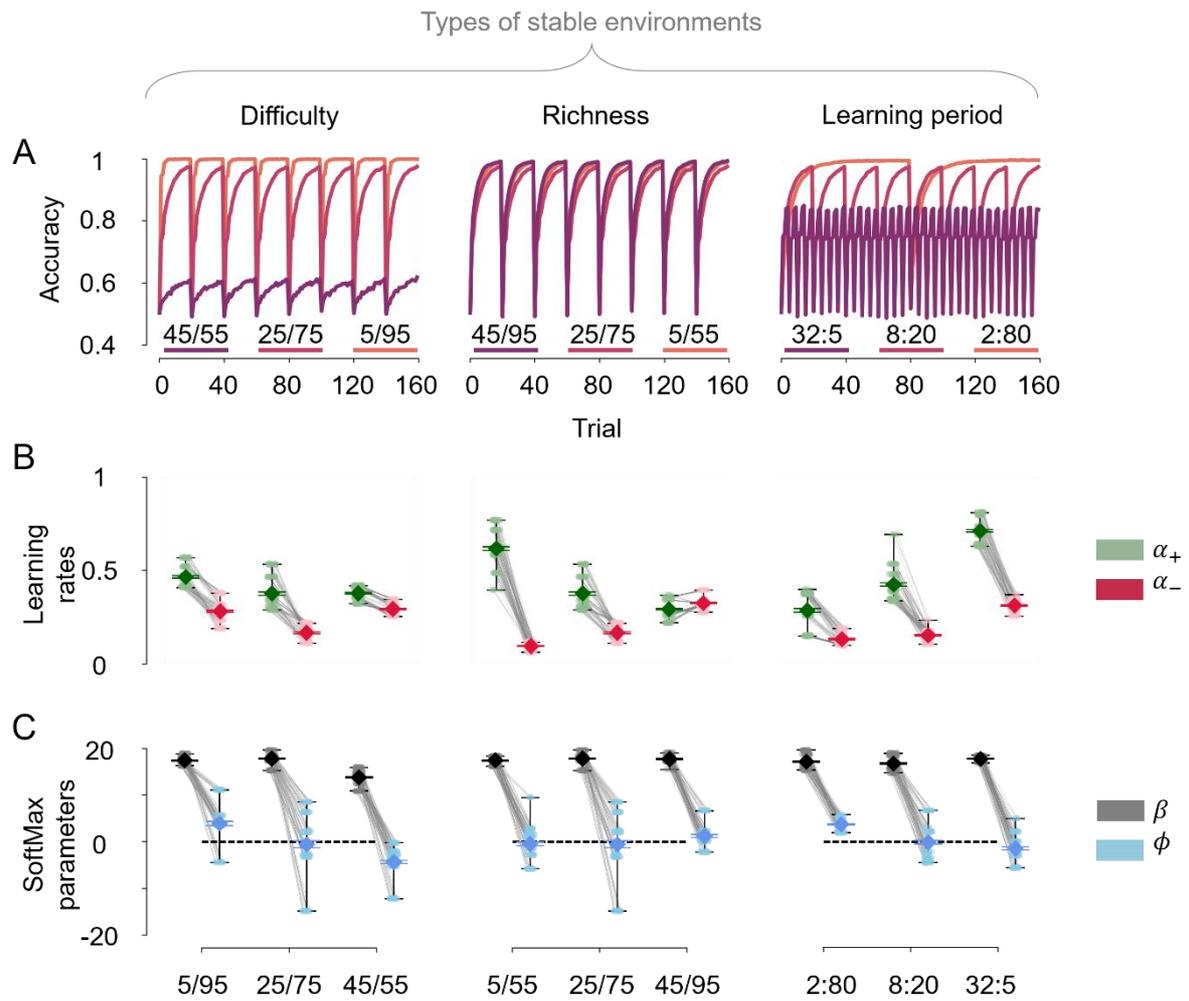

**Figure 4: effect of factor variation (difficulty, richness and learning period), with a uniform distribution of initial parameters.** (**A**) The top row represents the learning curves, averaged across agents and reboots, in each environment. (**B**) The middle row represents the learning rates ($\alpha_+$ in green and $\alpha_-$ in red, pairing in each reboot by a grey line). (**C**) The bottom row represents the paired values of the inverse temperature ($\beta$, in grey) perseveration ($\phi$, in blue). In (**B**) and (**C**) each grey line represents the results of one initialization of the whole evolutionary process (N=100).

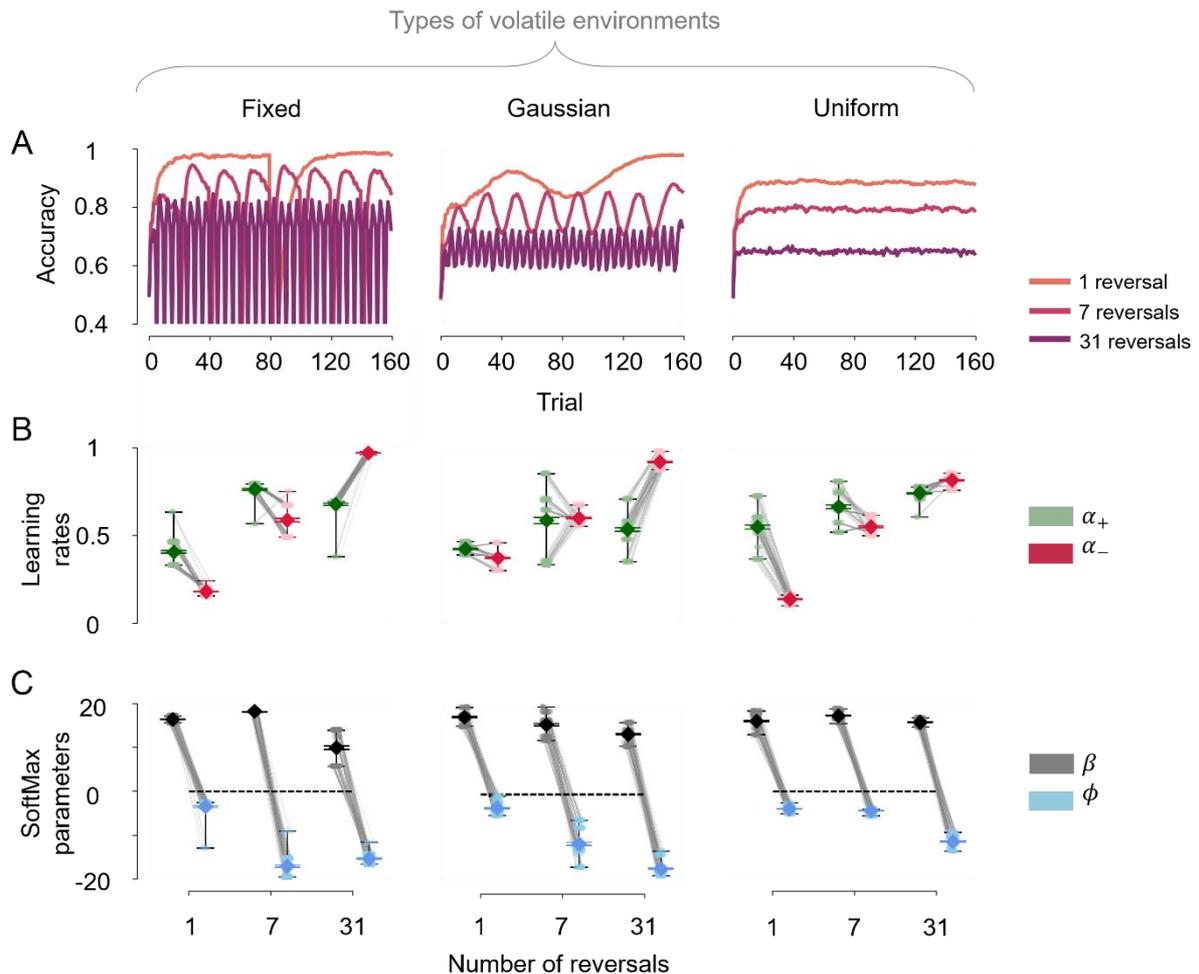

**Figure 5: effect of volatility (varying its probability distribution and frequency), with a uniform distribution of initial parameters.** (**A**) The top row represents the learning curves, averaged across agents and reboots, in each environment. (**B**) The middle row represents the learning rates ($\alpha_+$ in green and $\alpha_-$ in red, pairing in each reboot by a grey line). (**C**) The bottom row represents the paired values of the inverse temperature ($\beta$, in grey) perseveration ($\phi$, in blue). In (**B**) and (**C**) each grey line represents the results of one initialization of the whole evolutionary process (N=100).

When manipulating volatility, we noted that a positivity bias was generally selected in environments featuring a low and medium level of volatility and the opposite was observed in high volatility scenarios (negativity bias or pessimistic update[27,28]; **Figure 5**). We believe that this is because successful reversal in very volatile environments critically relies on the agent capacity to detect and consider negative feedback. Of note, we also observed a general increase of the average learning rate as a function the environment volatility as it has been theoretically postulated and observed[29,30].

We also observed a moderate-to-strong tendency for alternation ($\phi < 0$) was selected in all volatile environments. The observation that choice repetition ($\phi > 0$) was never selected in volatile environment (and in fact that the opposite vias seemed to be adaptive) reflects the fact that a tendency to alternate (i.e., swicht option) after many trials is adaptive in environments where such switches do occur.

**Evolution of parameters across stable versus volatile environments**

From these results (summarized on **Table 2; Figure 4** and **Figure 6)**, it seems that there is a dichotomy between the biases emerging in the stable and volatile environments. To further verify this point, we run simulations two 'macro' environments, encompassing respectively all the stable unique task specifications (7) and all the volatile unique task specifications (9), to observe the resulting overall results.

By analyzing these two sets of simulations, we observed that the learning rate for stable environments was lower than the one for volatile environments (in accordance with past works: high volatility environments require higher learning to quickly adapt to changes, while over-reacting to irregularities in stable environments is sub-optimal[29–31].

In the stable 'macro' environment, we observe the emergence of a positivity bias ($\Delta \alpha > 0$), as well as a tendency for perseveration ($\phi < 0$). In the volatile 'macro' environment, we also observed the emergence of a positivity bias, while the choice trace bias settled to a tendency for alternation ($\phi < 0$).

The emergence of a positivity bias in volatile environments may appear in contrasts with our previous observation that the negativity bias increases with high volatility. However, we implemented only three environments with very high volatility (average learning period of 5 choices), against six environments of low to medium volatility. Additionally, since the fitness is a function of accuracy and accuracy is higher in the low volatility environments, parameters that optimize performance in the low volatility environments will be prioritized by the selection algorithm.

Nonetheless, this result illustrate once again that the positivity bias is somehow more stably selected compared to gradual perseveration, as a source once again more stably evolved, as it is selected in both stable and volatile 'macro' environments, while perseveration is not. In fact, the opposite bias (gradual alternance; $\phi < 0$) seems to bear an adaptive value in volatile environments.

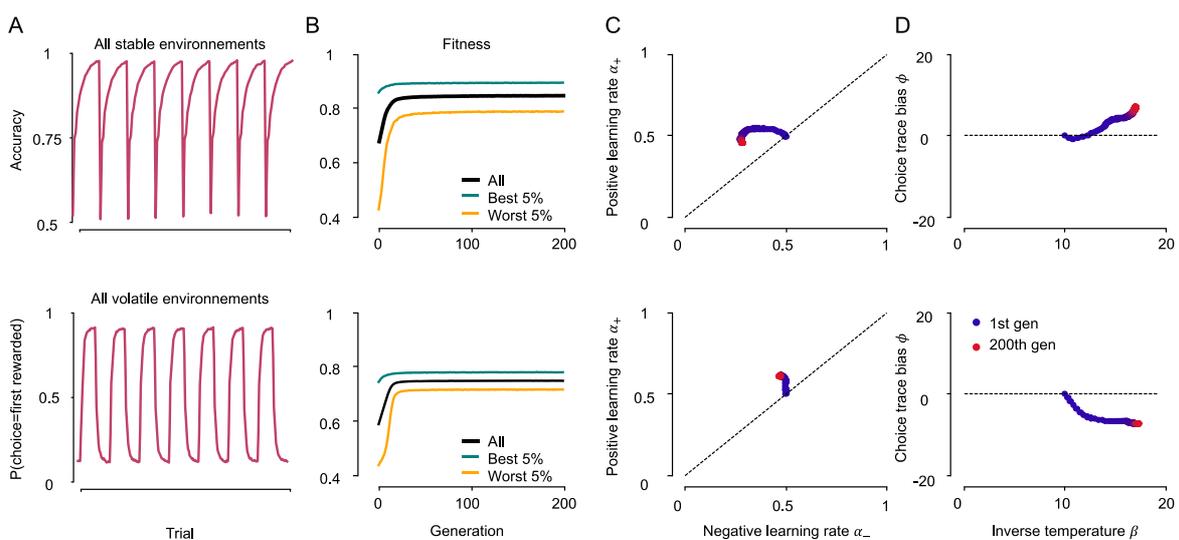

**Figure 6: global emergence of choice hysteresis biases in all stable (top row) and all volatile (bottom row) environments.** The top row represents the results obtained when agents faced all the possible stable environments (see Figure 4). The bottom row represents the results obtained when

agents faces all the volatile environments. (**A**) Evolution of the choice rate averaged across agents and reboots. (**B**) evolution of the fitness across generation averaged across reboots. (**C**) Evolution of the learning rates across generations averaged across reboots. (**D**) evolution of the SoftMax parameters across generation averaged across reboots).

**Table 2:** Summary of the results across scenarios. Each scenario was simulated 100 times (each simulation involving 1000 agents and 200 generations). *Difficulty*: probability of reward for the best/worst option. *Period*: Length of the learning period (number of repetitions). *Volatility*: the number of reversals (number of trials between reversals). $\Delta\alpha = \alpha_+ - \alpha_-$: update bias (>0 denotes positivity bias; <0 negativity). $\phi$: gradual perseveration bias (>0 denotes perseveration; <0 alternation).

|  |  |  | Variations | | |
|---|---|---|---|---|---|
| **Stable contingences** | ***Difficulty*** |  | *95/05* | *75/25* | *55/45* |
|  | Difference in reward probability | $\Delta\alpha > 0$ | 100% | 100% | 100% |
|  |  | $\phi > 0$ | 84% | 52% | 0% |
|  | ***Average reward*** |  | *5/55* | *25/75* | *45/95* |
|  | Average reward probability | $\Delta\alpha > 0$ | 100% | 100% | 36% |
|  |  | $\phi > 0$ | 52% | 52% | 80% |
|  | ***Learning period*** |  | *2(80)* | *8(20)* | *32(5)* |
|  | Duration and number of learning periods | $\Delta\alpha > 0$ | 100% | 100% | 100% |
|  |  | $\phi > 0$ | 100% | 36% | 36% |
| **Variable Contingences** | ***Fixed*** |  | *1 reversal* | *7 reversals* | *31 reversals* |
|  | Update | $\Delta\alpha > 0$ | 100% | 96% | 0% |
|  | Decision | $\phi > 0$ | 0% | 0% | 0% |
|  | ***Gaussian*** |  | *1 reversal* | *7 reversals* | *31 reversals* |
|  | Update | $\Delta\alpha > 0$ | 84% | 52% | 0% |
|  | Decision | $\phi > 0$ | 0% | 0% | 0% |
|  | ***Uniform*** |  | *1 reversal* | *7 reversals* | *31 reversals* |
|  | Update | $\Delta\alpha > 0$ | 100% | 84% | 0% |
|  | Decision | $\phi > 0$ | 0% | 0% | 0% |

**Table 3**: average bias values and significance (one-sample $t$-test against zero across reboots for $\alpha_+ - \alpha_-$ and $\phi$. Values in black indicate non-significant differences ($p > 0.05$), values in green indicate significantly positive values (i.e. a positivity bias and perseveration respectively, $p < 0.001$), and red values indicate significantly negative values (i.e. a pessimism bias and alternation). Significant differences are in bold.

|  |  |  | Variations | | |
|---|---|---|---|---|---|
| **Stable contingences** | ***Difficulty*** |  | *95/05* | *75/25* | *55/45* |
|  | Difference in reward probability | $\Delta\alpha$ | $0.18 \pm 0.006$ | $0.21 \pm 0.009$ | $0.08 \pm 0.002$ |
|  |  | $\phi$ | $3.97 \pm 0.45$ | $-0.49 \pm 0.76$ | $-4.29 \pm 0.37$ |
|  | ***Average reward*** |  | *5/55* | *25/75* | *45/95* |
|  | Average reward probability | $\Delta\alpha$ | $0.52 \pm 0.009$ | $0.21 \pm 0.009$ | $-0.03 \pm 0.004$ |
|  |  | $\phi$ | $-0.38 \pm 0.35$ | $-0.49 \pm 0.76$ | $1.27 \pm 0.27$ |
|  | ***Learning period*** |  | *2(80)* | *8(20)* | *32(5)* |
|  | Duration and number of learning periods | $\Delta\alpha$ | $0.15 \pm 0.008$ | $0.27 \pm 0.008$ | $0.4 \pm 0.005$ |
|  |  | $\phi$ | $3.64 \pm 0.13$ | $-0.16 \pm 0.37$ | $-1.47 \pm 0.27$ |
| **Variable Contingences** | ***Fixed*** |  | *1 reversal* | *7 reversals* | *31 reversals* |
|  | Update | $\Delta\alpha$ | $0.22 \pm 0.009$ | $0.17 \pm 0.009$ | $-0.29 \pm 0.005$ |
|  | Decision | $\phi$ | $-3.39 \pm 0.20$ | $-17.02 \pm 0.26$ | $-15.29 \pm 0.13$ |
|  | ***Gaussian*** |  | *1 reversal* | *7 reversals* | *31 reversals* |
|  | Update | $\Delta\alpha$ | $0.05 \pm 0.004$ | $-0.01 \pm 0.02$ | $-0.38 \pm 0.008$ |
|  | Decision | $\phi$ | $-3.1 \pm 0.15$ | $-11.33 \pm 0.35$ | $-17.00 \pm 0.17$ |
|  | ***Uniform*** |  | *1 reversal* | *7 reversals* | *31 reversals* |
|  | Update | $\Delta\alpha$ | $0.41 \pm 0.09$ | $0.11 \pm 0.01$ | $-0.07 \pm 0.005$ |
|  | Decision | $\phi$ | $-3.9 \pm 0.06$ | $-4.39 \pm 0.05$ | $-11.39 \pm 0.15$ |

**Discussion**

Positivity bias and gradual perseveration have been both put forward as robust features of human reinforcement learning in bandit tasks[2,3]. These computational processes are often framed as biases, as they are shown to increase choice repetition or hysteresis above and beyond what would be expected from the actual history of past outcomes[6]. Despite their well-established empirical foundation, comparatively little systematic investigations were devoted to understanding their possible advantages (normative foundation).

To this purpose, we implemented an evolutionary approach where populations of agent competed in terms of performance in various bandit tasks. By doing so we followed the footsteps of many previous studies applied evolutionary technique to reinforcement learning algorithms in order to speed up the discovery of optimal solutions[32]. The agents were represented by cognitive model of reinforcement learning adapted for bandit tasks and inspired by the model fitted from human behavior in similar contexts[3]. More specifically the agent's genotype was represented by five free parameters, which, under certain combinations, could generative positive/negativity biases and gradual perseveration/alternance.

Basic sanity checks illustrated that our algorithm worked, notably the fitness function was well-behaved (monotonically increasing and negatively accelerated)[33] and the results were stable across multiple initializations and variants of the algorithm. Furthermore, orthogonally to positivity bias and gradual perseveration, our findings recapitulated known principles, notably concerning the relation between the average learning rate and the volatility. Indeed, all our results confirmed that increasing the volatility of the environment lead to the development of higher learning rates as it has been previously empirically and normatively demonstrated[29–31,34].

Concerning the evolution of learning rates, we found that it was quasi-systematically selected in the stable environments and selected in most the volatile environments (with the notable exception of the environment with higher volatility). Our results therefore indicate that positivity bias gives a computational advantage over unbiased update leading to higher performance, average. We believe that the positivity bias is beneficial because it implements a form of *noise cancellation*. The positivity bias will tend to over-estimate the value of the high-reward options and under-estimate the value of low-reward options. Such a bias thus enhances the discrepancy between two options, therefore making it more robust to the stochasticity of the reward[14,15].

Our results are overall consistent and extend a previous study showing that an asymmetric updated akin to positivity bias (confirmation bias[35]) is also generally optimal in bandit tasks featuring complete feedback (i.e., simultaneous information about the chosen and unchosen options' outcomes[15,36]). By showing that the positivity bias also emerges in tasks with partial feedback, illustrates that its advantages are also maintained in situations where there is a tension between exploration and exploitation (which is absent in complete feedback scenarios).

Our results are also consistent with a recent study, showing that positivity bias emerge as a feature of a neural network trained through meta-reinforcement learning[37] to optimally solve bandit task, whose stable contingencies were procedurally generated in a space overlapping with those of the environments included here[38]. The convergence between our and their results is consistent with the idea that neural network training is better understood as a metaphor of evolutionary processes rather experiential learning[39].

Among the simulations of tasks with stable contingencies, the "rich" environment was the only one which selected a negativity bias. These results is consistent a previous study where Cazé and van der Meer[14] have shown that the update asymmetry is systematically beneficial in high- and low-reward conditions, with positivity bias enhancing correct decisions in low-reward environments and negativity bias being more adaptive in high-reward environments. The reason for this asymmetry is that in rich environments, both options yield a high reward, so a higher contrast can be achieved through neglecting positive outcomes (i.e., a negative bias in the rich environment correspond to "paying more attention" to the rare outcomes).

In volatile environments, we observed that the direction of the asymmetric update depends on the frequency of the reversals, with more frequent reversals resulting in selecting a negativity bias. an agent needs to adapt quickly in highly volatile environments, i.e. it needs to be sensitive to when a reversal occurs. After learning to choose the correct option in a volatile environment, a negative outcome likely indicates that the option values were reversed. Increasing the behavioral sensitivity to negative outcomes (negativity bias) therefore encourages the fast adaptation to contingency reversal.

Intriguingly, studies explored the role of learning biases beyond in individual learning, in social learning and also found beneficial effects. Bergerot et al.[40] showed that, in the context of multi-agent learning, a positivity bias allow large groups of virtual agents to perform better than unbiased agents. However, a putative positive role of positivity bias in social learning, is mitigated by other observations suggesting that similar biases can produce to polarization and other undesirable outcomes[41,42].

Our study also shows that gradual perseveration is less systematically selected, even in stable environments, where it emerges only in easy and rich environments, or with long learning periods. In principle, gradual perseveration, by increasing choice rate for a given option, irrespective of the history of outcomes, can also increase performance by neglecting stochastic negative outcomes (noise cancellation). However, gradual perseveration will exert a positive effect, only after that the correct response has been identified and before that point it could be detrimental. Therefore, it is unsurprising that it is selected in long learning periods and easy tasks.

In contrast, we noted that gradual alternation it was systematically selected in volatile environments. The reason why perseveration is not selected in volatile environments is easily derived from the fact that gradual perseveration undermines the agents' capacity to react to feedback and switch responses. In fact, in these environments the opposite is selected, indicating that gradual alternance (that could be understood as the building up of a progressive "urge" to change option, regardless of the past outcomes) is useful to anticipate the reversals. This interpretation is confirmed by the fact that the tendency for alternation is proportional with the frequency of reversals: once an alternative has been repeated enough times, its weight in the choice probability is such that it induces a switch (ideally in concomitance with a reversal).

To sum up, our study showed that positivity bias generally seems to be a desirable property for agent solving a variety of bandit, while the optima sign of choice history-dependence seems to be less robustly selected and strongly affected by the volatility of the environment. Overall, we also observed that across environs, the value of selected parameter values changed greatly. This suggests that the optimal parameters are largely environment-specific, and our findings can be leveraged by future research to design meta-reinforcement learning agents whose parameters adapt to the statistics and the structure of the environment in a manner that is coherent with our results[43–45].

In conclusion, our findings provide new insights normative foundations of positivity bias and gradual perseveration in reinforcement learning. By illustrating the comparative advantages of these processes, our work has the potential to inform and shape future research in psychology, neuroscience, and artificial intelligence. Broadly speaking our results contribute reinforcing the idea that, contrary to the traditional point of view, cognitive biases can have positive value and increased performance[22].

# Methods

*The bandit task*

Virtual agents had to perform a two-armed bandit task with partial feedback. Each option is associated with a fixed probability of obtaining a reward. At each trial, the agents would either get a reward of value +1 or a punishment of value -1.

*Model parametrization*

The parameters are bounded such that $\alpha_+, \alpha_-, \tau$ are comprised between 0 and 1, $\beta$ between 0 and 20 and $\phi$ between -20 and 20. **Table 4** summarizes their possible values.

**Table 4 parameter definition and their respective mutation distributions.**

| Parameter | Value range | Mutation distribution |
|---|---|---|
| $\beta$ | [0,20] | $\mathcal{N}(0,1)$ |
| $\alpha_+$ | [0,1] | $\mathcal{N}(0,0.05)$ |
| $\alpha_-$ | [0,1] | $\mathcal{N}(0,0.05)$ |
| $\tau$ | [0,1] | $\mathcal{N}(0,0.05)$ |
| $\phi$ | $[-20, 20]$ | $\mathcal{N}(0,2)$ |

*The evolutionary algorithm*

Classical approaches to obtaining optimal parameters consist of variations of grid search, which are computationally extensive. Here, we propose an evolutionary approach through which the optimal parameters are progressively selected.

At a given generation, 1000 virtual agents perform the task, and are then ordered according to their performance, computed as the choice rate on the high reward probability option over all trials. The 5% (i.e. 50 agents) of agents with the lower performance are removed from the simulation, while the top 5% of performers are duplicated, hence maintaining 1000 agents in all generations. This process is repeated for 200 generations.

Phenotypical diversity is implemented in two ways. The first option is to start from a random assignment of parameter values, drawn from a uniform distribution, to each agent. Evolution then narrows down the span of the parameter space by progressively discarding sub-optimal sets of parameters.

The second option is to start from a common instantiation of all agents' parameters. At each generation and after the reproduction process, a small random mutation drawn from a Gaussian distribution is introduced to the parameters of 5% of all the surviving agents. All the mutation distributions are centered around zero, and their standard deviations are adapted to the range of each parameter, according to the formula:

$$\sigma_i = \frac{\text{range of parameter } i}{\text{range of } \beta}$$

(also see **Table 4** for the values). The parameters are then thresholded so they lie within the boundaries stated above.

To ensure the stability of the results and the robustness of the evolutionary algorithm, each experiment was repeated 100 times.

**Acknowledgments**

The authors thank Jean-Baptiste André for useful insights about evolutionary algorithms. SP is funded by the European Research Council consolidator grant (RaReMem: 101043804) and three Agence Nationale de la Recherche grants (CogFinAgent: ANR-21-CE23-0002-02; RELATIVE: ANR-21-CE37- 750 0008-01; RANGE: ANR-21-CE28-0024-01) and the Alexander Von Humbolt foundation.